\documentclass[journal]{IEEEtran}

\ifCLASSINFOpdf
\else
   \usepackage[dvips]{graphicx}
\fi
\usepackage{url}

\usepackage{graphicx}
\usepackage{amsmath}
\usepackage{booktabs}
\usepackage{multirow}
\usepackage{algorithm}
\usepackage{algpseudocode}
\usepackage{adjustbox}
\usepackage{amssymb}

\begin{document}

\title{On Aligning Hierarchical Standardized Embedding for Audio-visual Generalized Zero-shot Learning}

\author{Zihan Zhang$^\dag$, Jie Hong$^\dag$, Siyuan Fan, Yanghao Zhou, \IEEEmembership{Student Member, IEEE} and Pengfei Fang$^*$, \IEEEmembership{Member, IEEE}
\thanks{$\dag$: Contribute equally to this research.}
\thanks{$^*$: Corresponding author (e-mail: fangpengfei@seu.edu.cn).}
\thanks{Zihan Zhang and Siyuan Fan are with School of Software Engineering, Southeast University, Nanjing, 210096, China. Jie Hong is with Faculty of Engineering, The University of Hong Kong, Pok Fu Lam, 999077, Hong Kong SAR. Yanghao Zhou is with Beijing Institute of Technology, Beijing, 100811, China. Pengfei Fang is with Key Laboratory of New Generation Artificial Intelligence Technology and Its Interdisciplinary Applications (Southeast University), Ministry of Education, Nanjing, 210096, China. Pengfei Fang is also with School of Computer Science and Engineering, Southeast University, 210096, Nanjing, China.}}


\markboth{Journal of \LaTeX\ Class Files, Vol. 14, No. 8, August 2015}
{Shell \MakeLowercase{\textit{et al.}}: Bare Demo of IEEEtran.cls for IEEE Journals}
\maketitle

\begin{abstract}
Audio-visual Generalized Zero-shot Learning (AV-GZSL) is a challenging task that aims to classify both seen and unseen objects or scenes by integrating data from audio and visual modalities. Recent studies primarily focus on fusing or aligning audio and visual features to generate more informative audio-visual embeddings. Also, aligning the audio-visual and textual features of most existing methods relies solely on the optimization objectives. However, those methods neglect the inherent distributional and structural differences between audio-visual and textual modalities. To address this limitation, we propose a method termed Aligning Hierarchical Standardized Embedding (AHSE), which enables hierarchical alignment of standardized audio-visual and textual embeddings within a shared embedding space. Specifically, we first apply Z-score standardization to the fused audio-visual and textual embeddings to reduce distributional mismatches. We then introduce a hierarchical alignment strategy that minimizes discrepancies at the semantic, class, and batch levels, thereby constructing a more robust and well-structured embedding space. This strategy not only preserves semantic and inter-class relationships but also maintains spatial consistency within each batch. Extensive experiments on three benchmark datasets: VGGSound-GZSL, UCF-GZSL, and ActivityNet-GZSL, demonstrate that AHSE achieves competitive performance in zero-shot learning.
\end{abstract}


\IEEEpeerreviewmaketitle

\section{Introduction} 
\label{sec:intro}
In real-world scenarios, humans naturally integrate information from multiple sensory modalities—such as vision and hearing—to perform everyday activities. For instance, referee’s whistle, and reacts to the crowd’s cheers. Inspired by this ability, deep learning research has shown that models can benefit from multi-modal data, which provides richer and more diverse cues and thereby enhances their capacity for accurate understanding~\cite{cheng2020look,afouras2022self,ebenezer2021detection,xiao2020audiovisual,mercea2023text,zhou2024advancing,zhou2025aloha,zhang2025audio,xuan2025x}. Nonetheless, collecting large-scale multi-modal datasets remains a formidable challenge, as it requires significant time and financial resources. Moreover, unlike humans, deep learning models often struggle to recognize or predict samples from previously unseen classes in real-world environments.

To address this limitation, the task of Audio-Visual Generalized Zero-Shot Learning (AV-GZSL) has been introduced~\cite{parida2020coordinated}. AV-GZSL aims to classify objects or scenes by jointly exploiting audio and visual modalities, even in the absence of audio-visual training data from unseen classes. This task remains highly challenging, as it demands not only the effective fusion of audio and visual features but also the construction of a shared embedding space that aligns audio-visual representations with corresponding textual descriptions~\cite{parida2020coordinated,mazumder2021avgzslnet,mercea2022audio,mercea2022temporal,hong2023hyperbolic,li2023motion,zheng2023generative,li2024spiking,kurzendorfer2024audio,mo2025audio,yu2025discrepancy,ma2026fusion}.

Most existing AV-GZSL methods primarily emphasize fusing audio and visual features to obtain more informative multimodal embeddings. In parallel, conventional alignment strategies typically rely on straightforward optimization objectives such as cross-entropy or triplet losses within a shared embedding space~\cite{mercea2022audio,hong2023hyperbolic,li2024spiking}. However, these approaches often overlook the inherent distributional discrepancies between audio-visual and textual modalities, thereby limiting their ability to capture joint distributions effectively. Furthermore, traditional alignment methods underutilize the structural information contained within each modality. In particular, structural similarities across semantic levels between audio-visual and textual modalities can play a crucial role in determining alignment quality and thus warrant explicit modeling.

This paper introduces Aligning Hierarchical Standardized Embedding (AHSE), a method that simultaneously reduces distributional discrepancies and enhances structural alignment between audio-visual and textual modalities for the AV-GZSL task. Specifically, AHSE employs a distribution standardization strategy implemented via Z-score normalization to map the distributions of audio-visual and textual features onto a common Gaussian distribution. This process mitigates the hubness problem frequently observed in zero-shot learning~\cite{fei2021z}, while also alleviating modality-specific distributional differences. Beyond harmonizing distributions, Z-score standardization produces a more uniform embedding space, thereby improving feature separability and further reducing hubness. To complement this, AHSE introduces a hierarchical alignment strategy for the standardized embeddings. We design a novel loss function that imposes constraints on local structures across three levels—semantic, class, and batch. Optimizing this objective encourages the model to learn a shared embedding space that preserves semantic consistency while strengthening cross-modal alignment. It is noted that the proposed AHSE is built on the baseline CLIP-CLAP~\cite{kurzendorfer2024audio}. CLIP-CLAP is the foundational model for the AV-GZSL task and achieves SOTA performance compared to existing methods. The contributions of this work can be summarized as follows:
\begin{itemize}
    \item We propose AHSE, a novel approach to the AV-GZSL problem that jointly optimizes both distributional consistency and structural alignment between audio-visual and textual modalities.
    \item The proposed method integrates a distribution standardization strategy with a hierarchical alignment mechanism, effectively mitigating distributional discrepancies and hubness issues, while simultaneously enhancing the alignment of local structural relationships between audio-visual and textual modalities.
    \item We conduct extensive experiments across multiple benchmark datasets, demonstrating that our method effectively aligns audio-visual and textual features and delivers consistently competitive performance across all datasets.
\end{itemize}
\section{Method} \label{sec:method}
In this section, we first present the baseline model, CLIP-CLAP~\cite{kurzendorfer2024audio}, and then provide a detailed description of our proposed framework, Aligning Hierarchical Standardized Embedding (AHSE). The proposed method standardizes feature embeddings within a shared space and introduces a hierarchical alignment strategy that aligns audio-visual and textual embeddings across semantic, class, and batch levels. The overall architecture of AHSE is illustrated in Fig.~\ref{fig:framework}.

\subsection{Problem Formulation}
Audio-visual generalized zero-shot learning (AV-GZSL) is required to recognize audio-visual samples from both seen (S) and unseen (U) classes. In this setting, the training set that belongs to the seen classes can be represented as $\boldsymbol{S} =(\boldsymbol{v}_i^s,\boldsymbol{a}_i^s,\boldsymbol{w}_i^s,\boldsymbol{y}_i^s)_{i\in\{1,\ldots, N\}}$. In $\boldsymbol{S}$, $\boldsymbol{v}_i^s$ and $\boldsymbol{a}_i^s$ represent the visual and the audio features. $\boldsymbol{y}_i^s$ is the ground truth class label for sample $i$, and $\boldsymbol{w}_i^s$ is the textual description corresponding to the ground truth class label. $N$ is the number of samples in $\boldsymbol{S}$. Similarly, the set of samples from unseen classes consisting of $M$ elements is defined as $\boldsymbol{U}=(\boldsymbol{v}_{i}^{u},\boldsymbol{a}_{i}^{u},\boldsymbol{w}_{i}^{u},\boldsymbol{y}_{i}^{u})_{i\in\{1,\ldots, M\}}$. For $\boldsymbol{S}$ and $\boldsymbol{U}$, it is satisfying that $\boldsymbol{y}^s \cap \boldsymbol{y}^u = \emptyset$. The total number of classes is denoted as $K$, and the class label is given by $p\in\{1, 2, \ldots, K\}$. The numbers of seen and unseen classes are denoted by $K_s$ and $K_u$, respectively. The goal of AV-GZSL is to learn a common embedding space using seen samples (\textit{e.g.}, S) and apply it to recognize unseen samples (\textit{e.g.}, U). 

\subsection{Baseline}
\begin{figure}[t]
    \centering
    \includegraphics[width=0.47\textwidth]{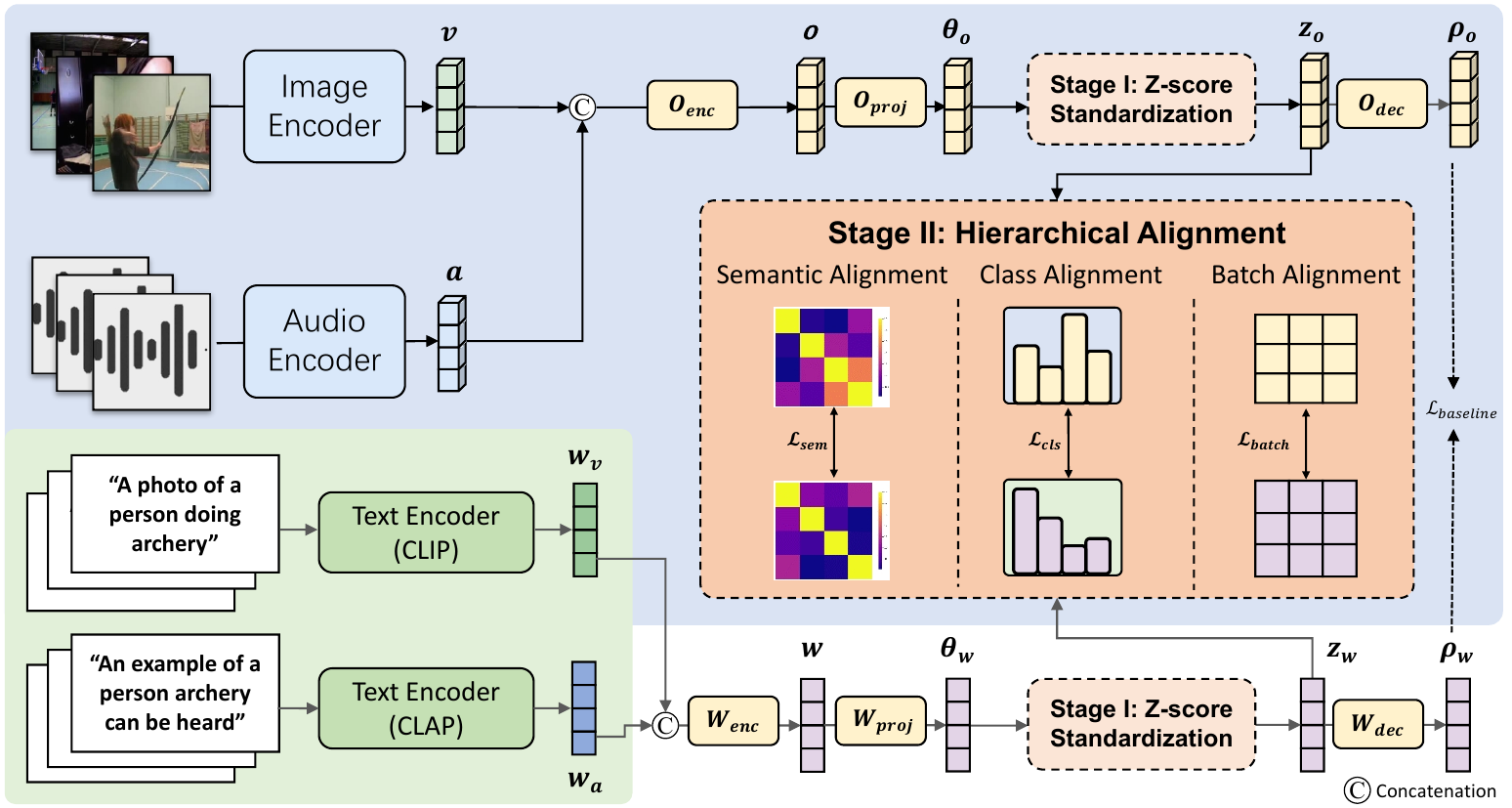}
    \vspace{-6pt}
    \caption{Overview of the proposed AHSE. Visual and audio features are extracted from raw data using the image and audio encoders from CLIP and CLAP, and then passed through multiple fully connected layers after concatenation to obtain audio-visual embeddings. Textual embeddings can be obtained similarly by text encoders from CLIP and CLAP. We apply Z-score standardization to both audio-visual and textual embeddings, then align the two standardized embeddings hierarchically.}
    \vspace{-12pt}
    \label{fig:framework}
\end{figure}

Most AV-GZSL methods typically follow a two-stage framework: first, they fuse audio and visual features to generate audio-visual embeddings, and second, construct a shared embedding space to align these embeddings with textual embeddings. Our method, AHSE, follows the common learning paradigm of AV-GZSL and is built upon the CLIP-CLAP baseline~\cite{kurzendorfer2024audio}. As illustrated in Fig.~\ref{fig:framework}, visual and audio inputs are denoted by $\boldsymbol{v}$ and $\boldsymbol{a}$, respectively, and textual features $\boldsymbol{w}_v$ and $\boldsymbol{w}_a$ provide semantic class descriptions. Specifically, $\boldsymbol{v}$ and $\boldsymbol{w}_v$ are extracted from the image encoder and text encoder of CLIP~\cite{radford2021learning}, whereas $\boldsymbol{a}$ and $\boldsymbol{w}_a$ are extracted from the audio encoder and text encoder from the CLAP~\cite{mei2024wavcaps}. To capture shared information across modalities, the fused audio-visual feature $\boldsymbol{o}$ is generated by passing the concatenation of $\boldsymbol{v}$ and $\boldsymbol{a}$ through the audio-visual encoder $O_{enc}$ (\textit{e.g.}, $\boldsymbol{o} = O_{enc}( \boldsymbol{v}  \copyright \boldsymbol{a} )$), while the fused textual feature $\boldsymbol{w}$ is obtained by feeding the concatenation of $\boldsymbol{w}_v$ and $\boldsymbol{w}_a$ into the word encoder $W_{enc}$ (\textit{e.g.}, $\boldsymbol{w} = W_{enc}( \boldsymbol{w}_v \copyright \boldsymbol{w}_a )$). These fused representations are then projected into a shared embedding space through projection networks $O_{proj}$ and $W_{proj}$, yielding the audio-visual embedding $\boldsymbol{\theta}_o$ and the textual embedding $\boldsymbol{\theta}_w$, respectively. Finally, to further enhance semantic consistency, decoder networks $O_{dec}$ and $W_{dec}$ reconstruct $\boldsymbol{\rho}_o$ and $\boldsymbol{\rho}_w$ from the embeddings, ensuring that the decoded outputs match the original textual feature $\boldsymbol{w}$.

\subsection{Aligning Hierarchical Standardized Embedding}
We propose a universal framework, called Aligning Hierarchical Standardized Embedding (AHSE), that jointly optimizes distributional and structural alignment between audio-visual and textual data by incorporating a distribution standardization mechanism and a hierarchical alignment strategy. This design aims to reduce distributional discrepancies, alleviate the hubness problem, and preserve local structural similarities between the two modalities.

\subsubsection{Stage I: Embedding Standardization.}
The effectiveness of AV-GZSL significantly depends on the alignment between audio-visual and textual embeddings. However, existing approaches often rely on aligning audio-visual and textual embeddings with simple optimization. This practice overlooks the inherent distributional differences between the two modalities in the embedding space. To address these distributional disparities, we standardize the audio-visual embedding $\boldsymbol{\theta}_o$ and textual embeddings $\boldsymbol{\theta}_w$ in the common space with Z-score standardization, which employs the mean and standard deviation, as follows: 
\begin{equation}
\mu(\boldsymbol{\theta}_o)=\frac{1}{n}\sum_{i}^{n}\boldsymbol{\theta}_{o_i}, \enspace
\sigma({\boldsymbol{\theta}_o})=\sqrt{\frac{1}{n}\sum_{i}^{n}(\boldsymbol{\theta}_{o_i}-\mu(\boldsymbol{\theta}_o))^{2}},
\label{eq:mean}
\end{equation}
where $\boldsymbol{z}_o = \frac{\boldsymbol{\theta}_o-\mu(\boldsymbol{\theta}_o)}{\sigma(\boldsymbol{\theta}_o)}$. $\boldsymbol{\theta}_{o_i}$ represents the audio-visual embedding for sample $i$ and $n$ denotes the number of training samples. The terms $\mu(\boldsymbol{\theta}_o)$ and $\sigma(\boldsymbol{\theta}_o)$ refer to the mean and standard deviation of the audio-visual embeddings, respectively. The term $\boldsymbol{z}_o$ refers to the audio-visual embedding after Z-score standardization. Similarly, we can obtain the textual embedding $\boldsymbol{z}_w$ following the same standardization process. This strategy effectively alleviates the impact of inherent distributional differences between audio-visual and textual embeddings and mitigates the hubness problem in zero-shot learning by reducing the intermodal discrepancy between audio-visual and text. 

\subsubsection{Stage II: Hierarchical Alignment.}
We hypothesize that there is structural similarity between the audio-visual and textual modalities, and that this similarity would influence their alignment.
To take advantage of this and improve performance, we propose a hierarchical alignment strategy to align standardized audio-visual and textual embeddings and to alleviate structural differences between the two modalities. The proposed hierarchical alignment comprises semantic, class, and batch structure alignments.

\noindent\textbf{Semantic Structure Alignment.}
In AV-GZSL, establishing a shared embedding space that effectively accommodates both audio-visual and textual embeddings is crucial. Within this space, each sample is represented as a vector. Ideally, vectors from different samples within the same class should exhibit uniform correlations across dimensions, a property we refer to as the semantic structure. This semantic structure should be consistent between audio-visual and textual embeddings to ensure meaningful alignment. The goal of semantic alignment is to minimize discrepancies in the semantic structures of the two modalities. 
We measure this structure using the outer product, where the semantic relationship between samples $i$ and $j$ is expressed as $\boldsymbol{z}_i^\top\cdot \boldsymbol{z}_j$. The overall semantic structure can be captured by a semantic correlation matrix $\boldsymbol{S}=\boldsymbol{z}^\top\cdot \boldsymbol{z}$, where $\boldsymbol{S}$ is an $m \times m$ matrix and $m$ denotes the dimensionality of the embeddings. Here, $\boldsymbol{z}$ represents the standardized audio-visual or textual embeddings in the shared space. The objective is to minimize the divergence between the semantic correlation matrices of the two modalities, and the corresponding semantic structure alignment loss is defined as follows:
\begin{equation}
    \mathcal{L}_{sem}=\frac{1}{m} ||\boldsymbol{S}_{w}-\boldsymbol{S}_{o}||_2^2,
\label{eq:semantic}
\end{equation}
where $\boldsymbol{S}_{w}$ and $\boldsymbol{S}_{o}$ are the semantic correlation matrix of textual and audio-visual embeddings.

\noindent\textbf{Class Structure Alignment.}
A key objective of AV-GZSL is to identify the textual embedding closest to the corresponding audio-visual embedding, where each textual embedding serves as a prototype for its class. Ideally, samples belonging to the same category should be closely clustered around their respective prototypes. Traditional methods typically achieve this by minimizing the Euclidean distance between embeddings. However, KL divergence offers a more detailed way to match distributions across modalities, going beyond simple distance-based approximations. By minimizing distributional differences through KL divergence, the audio-visual embeddings can better align with the structural properties of their corresponding textual embeddings, thereby increasing the likelihood of correct classification. Recent studies~\cite{chen2024tackling, zhou2025unialign} have also shown that KL divergence is effective in capturing both structural similarity and uncertainty across modalities.
We employ KL divergence for class structure alignment, and the corresponding loss can be expressed as:
\begin{equation}
    \mathcal{L}_{cls} = \frac{1}{n m} \left(\sum_{i, j} \boldsymbol{z}_{w_{i, j}} \log \boldsymbol{z}_{w_{i, j}}  - \sum_{i, j} \boldsymbol{z}_{w_{i, j}} \log \boldsymbol{z}_{o_{i, j}}\right),
\label{eq:class}
\end{equation}
where $\boldsymbol{z}_{o_{i,j}}$ and $\boldsymbol{z}_{w_{i,j}}$ indicate the value of the sample $i$ and the dimension $j$ of the standardized audio-visual embeddings and textual embeddings. $n$ is the number of training samples and $m$ is the number of dimensions for embeddings.

\noindent\textbf{Batch Structure Alignment.}
Inspired by RKD~\cite{park2019relational} and DKD~\cite{fang2024distilling}, we observe that spatial structures naturally exist among different samples within a batch in the embedding space. To leverage this property, we propose a batch alignment strategy that captures the spatial structural similarities between audio-visual and textual embeddings, encouraging different samples to share similar directions and distances. Our objective is for the audio-visual embeddings to effectively preserve the structural relationships derived from the corresponding textual embeddings within the same batch. We compute these relationships using inner products, where the batch structural relationship between samples $i$ and $j$ in the embedding space is represented as $\boldsymbol{z}_i\cdot \boldsymbol{z}_j^\top$. The overall structural relations within a batch are captured by a spatial correlation matrix $\boldsymbol{H}=\boldsymbol{z}\cdot \boldsymbol{z}^\top$, where $\boldsymbol{H}$ is an $n \times n$ matrix and $n$ is the number of training samples. Here, $\boldsymbol{z}$ denotes the standardized audio-visual or textual embeddings in the shared space.
The goal is to minimize the divergence between these similarities between audio-visual embeddings and textual embeddings, leading to the following batch structure alignment loss:
\begin{equation}
    \mathcal{L}_{batch}=\frac{1}{n} ||\boldsymbol{H}_{w}-\boldsymbol{H}_{o}||_2^2,
\label{eq:batch}
\end{equation}
where $\boldsymbol{H}_{w}$ and $\boldsymbol{H}_{o}$ are the spatial correlation matrices of textual and audio-visual embeddings. 

\subsection{Training Loss}
We collectively refer to the semantic structure alignment loss, class structure alignment loss, and batch structure alignment loss mentioned above as the alignment loss, given by:
\begin{equation}
    \mathcal{L}_{align}=\mathcal{L}_{sem} + \mathcal{L}_{cls} + \mathcal{L}_{batch}.
\label{eq:align}
\end{equation}

The resulting alignment loss is derived hierarchically for a batch of data, and minimizing it minimizes discrepancies between the two modalities. 

We denote the baseline loss function as $\mathcal{L}_{baseline}$, which, in the case of CLIP-CLAP~\cite{kurzendorfer2024audio}, consists of a cross-entropy loss $\mathcal{L}_{ce}$, a reconstruction loss $\mathcal{L}_{rec}$, and a regression loss $\mathcal{L}_{reg}$, which is given by: 
\begin{equation}
    \mathcal{L}_{baseline} = \mathcal{L}_{ce} + \mathcal{L}_{rec} + \mathcal{L}_{reg}.
\label{eq:base}\end{equation}

Building on this, the final objective of our proposed method incorporates the baseline loss along with alignment loss:
\begin{equation}
    \mathcal{L} = \mathcal{L}_{baseline} + \lambda \mathcal{L}_{align},
\label{eq:total}
\end{equation}
where $\lambda$ represents the weight of $\mathcal{L}_{align}$.

\section{Experiments}
In this section, we present extensive experiments to validate the effectiveness of our proposed AHSE.

\subsection{Quantitative Results.}
As shown in Tab.~\ref{tab:comparison}, our method achieves the highest harmonic mean (HM) among existing audio-visual generalized zero-shot learning approaches. On UCF-GZSL, our model attains an HM of 67.26\% and a ZSL accuracy of 56.00\%, surpassing the previous best results of 60.97\% and 50.92\%, respectively, by 6.29\% and 5.08\%. On ActivityNet-GZSL, our approach achieves HM and ZSL scores of 31.56\% and 25.81\%, respectively, outperforming the prior state-of-the-art scores of 27.95\% and 25.20\% by 3.61\% and 0.61\%, respectively. On VGGSound-GZSL, our model achieves HM and ZSL scores of 18.62\% and 13.03\%, respectively, comparable to EZ-AVOOD’s 18.21\% and 13.28\%. Moreover, our method consistently yields higher unseen-class accuracy (U) across all datasets, demonstrating the effectiveness of standardized embeddings in enhancing generalization. These results highlight the advantages of our approach, confirming that embedding standardization combined with hierarchical alignment within a shared space substantially improves performance in generalized audio-visual zero-shot recognition.

\begin{table}[h!]
    \vspace{-6pt}
    \caption{Performance of our model compared to SOTA AV-GZSL methods on the VGGSound-GZSL, UCF-GZSL, and ActivityNet-GZSL datasets. All the models are trained and evaluated using both CLIP and CLAP features.}
    \centering
    \resizebox{0.48\textwidth}{!}{
        \begin{tabular}{l|cccc|cccc|cccc}
            \toprule
            \multirow{2}{*}{\textbf{Method}} & 
            \multicolumn{4}{c|}{\textbf{VGGSound-GZSL}} & 
            \multicolumn{4}{c|}{\textbf{UCF-GZSL}} & 
            \multicolumn{4}{c}{\textbf{ActivityNet-GZSL}} \\
            & S & U & HM & ZSL & S & U & HM & ZSL & S & U & HM & ZSL \\
            \midrule
            CJME~\cite{parida2020coordinated} & 11.96 & 5.41 & 7.65 & 6.84 & 48.18 & 17.78 & 25.87 & 20.46 & 16.06 & 9.13 & 11.64 & 9.92 \\
            AVGZSLNet~\cite{mazumder2021avgzslnet} & 13.02 & 2.88 & 4.71 & 5.44 & 56.26 & 34.37 & 42.67 & 35.66 & 14.81 & 11.11 & 12.70 & 12.39 \\
            AVCA~\cite{mercea2022audio} & 32.47 & 6.81 & 10.56 & 8.16 & 34.90 & 38.67 & 36.69 & 38.67 & 24.04 & 19.88 & 21.76 & 20.88 \\
            Hyper-multiple~\cite{hong2023hyperbolic} & 21.99 & 8.12 & 11.87 & 8.47 & 43.52 & 39.77 & 41.56 & 40.28 & 20.52 & 21.30 & 20.90 & 22.18 \\
            CLIP-CLAP~\cite{kurzendorfer2024audio} & 29.68 & 11.12 & 16.18 & 11.53 & 77.14 & 43.91 & 55.97 & 46.96 & \textbf{45.98} & 20.06 & 27.93 & 22.76 \\
            EZ-AVOOD~\cite{liu2025extremely} & \textbf{39.33} & 11.84 & 18.21 & \textbf{13.28} & 83.53 & 48.01 & 60.97 & 50.92 & 41.56 & 21.06 & 27.95 & 25.20 \\
            \midrule
            AHSE (ours) & 35.23 & \textbf{12.66} & \textbf{18.62} & 13.03 & \textbf{88.05} & \textbf{54.41} & \textbf{67.26} & \textbf{56.00} & 45.29 & \textbf{24.22} & \textbf{31.56} & \textbf{25.81} \\
            \bottomrule
        \end{tabular}
    }
    \vspace{-6pt}
    \label{tab:comparison}
\end{table}

\subsection{Ablation Study}
\noindent \textbf{Impact of Different Components.}
To evaluate the individual contributions of the proposed Hierarchical Alignment (HA) and Embedding Standardization (ES) modules within the AHSE framework, we conduct an ablation study on three benchmark datasets, as reported in Tab.~\ref{tab:ab2}. The results clearly demonstrate the substantial positive impact of both components. Removing either HA or ES individually results in noticeable performance degradation compared to the full AHSE model, whereas removing both simultaneously leads to a severe drop in performance, particularly in the harmonic mean (HM). These findings confirm that HA effectively mitigates structural discrepancies between modalities, whereas ES alleviates distributional differences between audio-visual and textual embeddings. Together, they enable more robust cross-modal alignment and improved knowledge transfer from seen to unseen classes.

\subsection{t-SNE Visualization}
\begin{figure}[t!]
    \centering
    \includegraphics[width=1.0\linewidth]{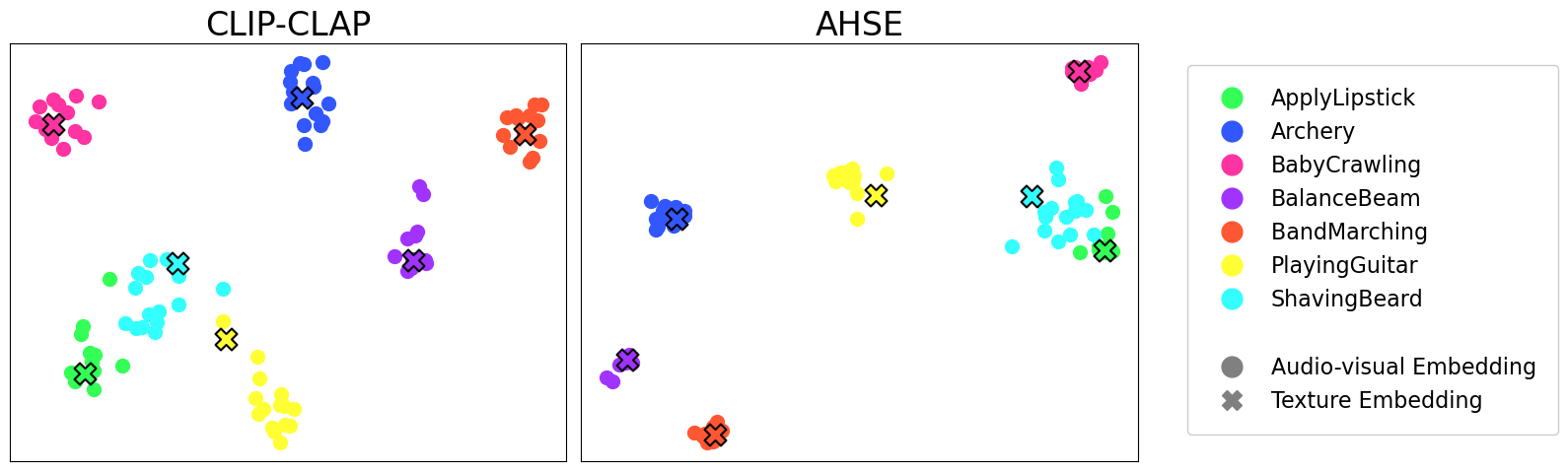}
    \caption{t-SNE visualization of CLIP-CLAP model and the proposed AHSE. We select 7 classes, of which ``PlayingGuitar'' and ``ShavingBeard'' are unseen, and all others are seen.}
    \vspace{-6pt}
    \label{fig:t-SNE}
\end{figure}

To visually assess the effectiveness of our AHSE framework for AV-GZSL, we present t-SNE visualizations on the UCF-GZSL dataset, comparing AHSE with the CLIP-CLAP baseline (see Fig.~\ref{fig:t-SNE}). The results show that sample points under AHSE form clearer clusters and exhibit stronger alignment than those produced by CLIP-CLAP, underscoring the benefits of addressing both distributional and structural discrepancies. For example, the separation between the unseen classes PlayingGuitar and ShavingBeard is noticeably improved, while instances within PlayingGuitar form more compact clusters, indicating more discriminative and well-structured embeddings.

\begin{table}[t!]
    \caption{Ablation study with different component combinations. ``w/o HA'' and ``w/o ES'' represent models without hierarchical alignment and embedding standardization.}
    \centering
    \resizebox{0.48\textwidth}{!}{
        \begin{tabular}{l|cc|cc|cc}
            \toprule
            \multirow{2}{*}{\textbf{Component}} & 
                \multicolumn{2}{c|}{\textbf{VGGSound-GZSL}} & 
                \multicolumn{2}{c|}{\textbf{UCF-GZSL}} & 
                \multicolumn{2}{c}{\textbf{ActivityNet-GZSL}} \\
            & HM & ZSL & HM & ZSL & HM & ZSL \\
            \midrule
            w/o HA \& ES & 16.18 & 11.53 & 55.97 & 46.96 & 27.93 & 22.76 \\
            w/o HA & 17.81 & 12.09 & 58.25 & 48.88 & 30.49 & 24.43 \\
            w/o ES & 18.61 & \textbf{13.41} & 61.81 & 51.57 & 30.39 & 25.50 \\
            AHSE & \textbf{18.62} & 13.03 & \textbf{67.26} & \textbf{56.00} & \textbf{31.56} & \textbf{25.81} \\
            \bottomrule
        \end{tabular}
    }
    \label{tab:ab2}
\end{table}

\section{Conclusion}
In this work, we studied the AV-GZSL problem and identified that the distributional and structural discrepancies between audio-visual and textual modalities have not been sufficiently addressed. To tackle this issue, we proposed AHSE, which first applies Z-score standardization to embeddings in the shared space to mitigate distributional differences and then employs a hierarchical alignment strategy to align audio-visual and textual representations across semantic, class, and batch structures. Extensive experiments on multiple benchmark datasets demonstrate that our method achieves consistently competitive results, highlighting both the effectiveness of standardized embeddings in reducing distributional mismatches and the advantages of hierarchical alignment in enhancing zero-shot performance. For future work, we plan to extend AHSE to a broader range of multimodal tasks that are more closely aligned with real-world applications.

\section*{Acknowledgment}
This work was supported by the National Natural Science Foundation of China (No. 62306070) and was also supported in part by the Southeast University Start-Up Grant for New Faculty under Grant 4009002309. Furthermore, the work was also supported by the Big Data Computing Center of Southeast University. This work was also supported by “the Fundamental Research Funds for the Central Universities (2242025K30024)”.

\bibliographystyle{IEEEtran}
\bibliography{bib.bib}

@inproceedings{parida2020coordinated,
  title={Coordinated joint multimodal embeddings for generalized audio-visual zero-shot classification and retrieval of videos},
  author={Parida, Kranti and Matiyali, Neeraj and Guha, Tanaya and Sharma, Gaurav},
  booktitle={Proceedings of the IEEE/CVF Winter Conference on Applications of Computer Vision},
  pages={3251--3260},
  year={2020}
}

@inproceedings{mazumder2021avgzslnet,
  title={Avgzslnet: Audio-visual generalized zero-shot learning by reconstructing label features from multi-modal embeddings},
  author={Mazumder, Pratik and Singh, Pravendra and Parida, Kranti Kumar and Namboodiri, Vinay P},
  booktitle={Proceedings of the IEEE/CVF Winter Conference on Applications of Computer Vision},
  pages={3090--3099},
  year={2021}
}

@inproceedings{mercea2022audio,
  title={Audio-visual generalised zero-shot learning with cross-modal attention and language},
  author={Mercea, Otniel-Bogdan and Riesch, Lukas and Koepke, A and Akata, Zeynep},
  booktitle={Proceedings of the IEEE/CVF Conference on Computer Vision and Pattern Recognition},
  pages={10553--10563},
  year={2022}
}

@inproceedings{mercea2022temporal,
  title={Temporal and cross-modal attention for audio-visual zero-shot learning},
  author={Mercea, Otniel-Bogdan and Hummel, Thomas and Koepke, A Sophia and Akata, Zeynep},
  booktitle={Proceedings of the European Conference on Computer Vision},
  pages={488--505},
  year={2022},
  organization={Springer}
}

@inproceedings{hong2023hyperbolic,
  title={Hyperbolic audio-visual zero-shot learning},
  author={Hong, Jie and Hayder, Zeeshan and Han, Junlin and Fang, Pengfei and Harandi, Mehrtash and Petersson, Lars},
  booktitle={Proceedings of the IEEE/CVF international Conference on Computer Vision},
  pages={7873--7883},
  year={2023}
}

@inproceedings{li2023motion,
  title={Motion-decoupled spiking transformer for audio-visual zero-shot learning},
  author={Li, Wenrui and Zhao, Xi-Le and Ma, Zhengyu and Wang, Xingtao and Fan, Xiaopeng and Tian, Yonghong},
  booktitle={Proceedings of the 31st ACM International Conference on Multimedia},
  pages={3994--4002},
  year={2023}
}

@inproceedings{kurzendorfer2024audio,
  title={Audio-Visual Generalized Zero-Shot Learning using Pre-Trained Large Multi-Modal Models},
  author={Kurzend{\"o}rfer, David and Mercea, Otniel-Bogdan and Koepke, A and Akata, Zeynep},
  booktitle={Proceedings of the IEEE/CVF Conference on Computer Vision and Pattern Recognition},
  pages={2627--2638},
  year={2024}
}

@article{li2024spiking,
  title={Spiking tucker fusion transformer for audio-visual zero-shot learning},
  author={Li, Wenrui and Wang, Penghong and Xiong, Ruiqin and Fan, Xiaopeng},
  journal={IEEE Transactions on Image Processing},
  year={2024},
  publisher={IEEE}
}

@inproceedings{mo2025audio,
  title={Audio-visual generalized zero-shot learning the easy way},
  author={Mo, Shentong and Morgado, Pedro},
  booktitle={Proceedings of the European Conference on Computer Vision},
  pages={377--395},
  year={2025},
  organization={Springer}
}

@article{fang2024distilling,
  title={On Distilling the Displacement Knowledge for Few-Shot Class-Incremental Learning},
  author={Fang, Pengfei and Qin, Yongchun and Xue, Hui},
  journal={arXiv preprint arXiv:2412.11017},
  year={2024}
}

@inproceedings{radford2021learning,
  title={Learning transferable visual models from natural language supervision},
  author={Radford, Alec and Kim, Jong Wook and Hallacy, Chris and Ramesh, Aditya and Goh, Gabriel and Agarwal, Sandhini and Sastry, Girish and Askell, Amanda and Mishkin, Pamela and Clark, Jack and others},
  booktitle={Proceedings of the International Conference on Machine Learning},
  pages={8748--8763},
  year={2021},
  organization={PMLR}
}

@article{mei2024wavcaps,
  title={Wavcaps: A chatgpt-assisted weakly-labelled audio captioning dataset for audio-language multimodal research},
  author={Mei, Xinhao and Meng, Chutong and Liu, Haohe and Kong, Qiuqiang and Ko, Tom and Zhao, Chengqi and Plumbley, Mark D and Zou, Yuexian and Wang, Wenwu},
  journal={IEEE/ACM Transactions on Audio, Speech, and Language Processing},
  year={2024},
  publisher={IEEE}
}

@inproceedings{park2019relational,
  title={Relational knowledge distillation},
  author={Park, Wonpyo and Kim, Dongju and Lu, Yan and Cho, Minsu},
  booktitle={Proceedings of the IEEE/CVF Conference on Computer Vision and Pattern Recognition},
  pages={3967--3976},
  year={2019}
}

@inproceedings{zheng2023generative,
  title={A generative approach to audio-visual generalized zero-shot learning: Combining contrastive and discriminative techniques},
  author={Zheng, Qichen and Hong, Jie and Farazi, Moshiur},
  booktitle={Proceedings of the International Joint Conference on Neural Networks},
  pages={1--8},
  year={2023},
  organization={IEEE}
}

@inproceedings{mercea2023text,
  title={Text-to-feature diffusion for audio-visual few-shot learning},
  author={Mercea, Otniel-Bogdan and Hummel, Thomas and Koepke, A Sophia and Akata, Zeynep},
  booktitle={Proceedings of the DAGM German Conference on Pattern Recognition},
  pages={491--507},
  year={2023},
  organization={Springer}
}

@article{xiao2020audiovisual,
  title={Audiovisual slowfast networks for video recognition},
  author={Xiao, Fanyi and Lee, Yong Jae and Grauman, Kristen and Malik, Jitendra and Feichtenhofer, Christoph},
  journal={arXiv preprint arXiv:2001.08740},
  year={2020}
}

@inproceedings{ebenezer2021detection,
  title={Detection of audio-video synchronization errors via event detection},
  author={Ebenezer, Joshua P and Wu, Yongjun and Wei, Hai and Sethuraman, Sriram and Liu, Zongyi},
  booktitle={Proceedings of the IEEE International Conference on Acoustics, Speech and Signal Processing},
  pages={4345--4349},
  year={2021},
  organization={IEEE}
}

@inproceedings{afouras2022self,
  title={Self-supervised object detection from audio-visual correspondence},
  author={Afouras, Triantafyllos and Asano, Yuki M and Fagan, Francois and Vedaldi, Andrea and Metze, Florian},
  booktitle={Proceedings of the IEEE/CVF Conference on Computer Vision and Pattern Recognition},
  pages={10575--10586},
  year={2022}
}

@inproceedings{cheng2020look,
  title={Look, listen, and attend: Co-attention network for self-supervised audio-visual representation learning},
  author={Cheng, Ying and Wang, Ruize and Pan, Zhihao and Feng, Rui and Zhang, Yuejie},
  booktitle={Proceedings of the 28th ACM International Conference on Multimedia},
  pages={3884--3892},
  year={2020}
}

@inproceedings{fei2021z,
  title={Z-score normalization, hubness, and few-shot learning},
  author={Fei, Nanyi and Gao, Yizhao and Lu, Zhiwu and Xiang, Tao},
  booktitle={Proceedings of the IEEE/CVF International Conference on Computer Vision},
  pages={142--151},
  year={2021}
}

@article{liu2025extremely,
  title={Extremely Simple Out-of-distribution Detection for Audio-visual Generalized Zero-shot Learning},
  author={Liu, Yang and Zhang, Xun and Du, Jiale and Gao, Xinbo and Han, Jungong},
  journal={arXiv preprint arXiv:2503.22197},
  year={2025}
}

@inproceedings{chen2024tackling,
  title={Tackling uncertain correspondences for multi-modal entity alignment},
  author={Chen, Liyi and Sun, Ying and Zhang, Shengzhe and Ye, Yuyang and Wu, Wei and Xiong, Hui},
  booktitle={The Thirty-eighth Annual Conference on Neural Information Processing Systems},
  year={2024}
}

@inproceedings{zhou2025unialign,
  title={UNIALIGN: Scaling Multimodal Alignment within One Unified Model},
  author={Zhou, Bo and Li, Liulei and Wang, Yujia and Liu, Huafeng and Yao, Yazhou and Wang, Wenguan},
  booktitle={Proceedings of the Computer Vision and Pattern Recognition Conference},
  pages={29644--29655},
  year={2025}
}

@inproceedings{yu2025discrepancy,
  title={Discrepancy-Aware Attention Network for Enhanced Audio-Visual Generalized Zero-Shot Learning},
  author={Yu, Runlin and Gong, Yipu and Li, Wenrui and Sun, Aiwen and Zheng, Mengren},
  booktitle={Proceedings of the 33rd ACM International Conference on Multimedia},
  pages={1112--1121},
  year={2025}
}

@article{ma2026fusion,
  title={Fusion-regularized alignment modality-adaptive audio-visual network for audio-visual zero-shot learning},
  author={Ma, Siteng and Niu, Xiaoyu and Tang, Haocheng and Yang, Zhe and Chu, Jisheng and Li, Wenrui},
  journal={Neurocomputing},
  pages={133693},
  year={2026},
  publisher={Elsevier}
}

@article{zhou2024advancing,
  title={Advancing weakly-supervised audio-visual video parsing via segment-wise pseudo labeling},
  author={Zhou, Jinxing and Guo, Dan and Zhong, Yiran and Wang, Meng},
  journal={International Journal of Computer Vision},
  volume={132},
  number={11},
  pages={5308--5329},
  year={2024},
  publisher={Springer}
}

@article{zhou2025aloha,
  title={Aloha: Adapting local spatio-temporal context to enhance the audio-visual semantic segmentation},
  author={Zhou, Yang-Hao and Huang, Heyan and Guo, Cunhan and Tu, Rong-Cheng and Xiao, Zeyu and Wang, Bo and Mao, Xian-Ling},
  journal={ACM Transactions on Multimedia Computing, Communications and Applications},
  volume={21},
  number={6},
  pages={1--23},
  year={2025},
  publisher={ACM New York, NY}
}

@article{zhang2025audio,
  title={Audio-Visual Event Localization with Cross Co-Attention and Dynamic Audio-Object Semantic Alignment},
  author={Zhang, Pufen and Shi, Peng and He, Xiao},
  journal={IEEE Signal Processing Letters},
  year={2025},
  publisher={IEEE}
}

@article{xuan2025x,
  title={X-STA: Cross-Modal Spatial-Temporal Alignment Network for Unified Audio-Visual Segmentation},
  author={Xuan, Hanyu and Liu, Tongxing and Dong, Wenxiang and Li, Zhongheng and Chen, Shuo},
  journal={IEEE Signal Processing Letters},
  year={2025},
  publisher={IEEE}
}

\end{document}